\documentclass{ieeeaccess}

\usepackage{cite}
\usepackage{amsmath,amssymb,amsfonts}
\usepackage{algorithmic}
\usepackage{graphicx}
\usepackage{textcomp}
\usepackage{tabularx}
\usepackage{algorithmic}
\usepackage{algorithm}
\usepackage{array}
\usepackage[caption=false,font=normalsize,labelfont=sf,textfont=sf]{subfig}
\usepackage{stfloats}
\usepackage{url}
\usepackage{verbatim}
\usepackage{xcolor}
\usepackage{hyperref} 
\usepackage{booktabs}
\usepackage{enumitem} 
\usepackage{caption}

\usepackage{pdflscape}
\usepackage{longtable}
\usepackage{adjustbox}
\usepackage{fancyhdr}

\def\BibTeX{{\rm B\kern-.05em{\sc i\kern-.025em b}\kern-.08em
    T\kern-.1667em\lower.7ex\hbox{E}\kern-.125emX}}

\begin{document}

\title{A Survey on Spoken Italian Datasets and Corpora}
\history{Date of publication xxxx 00, 0000, date of current version xxxx 00, 0000.}
\def\thevol{XX}
\def\theyear{202X}
\doi{XXX}
\author{
\uppercase{Marco Giordano (ORCID: 0009-0001-1649-6085)}\authorrefmark{1},
\uppercase{Claudia Rinaldi (ORCID: 0000-0002-1356-8151)}\authorrefmark{2}}

\address[1]{Department of Information Engineering, Computer Science and Mathematics, Università degli Studi dell'Aquila (e-mail: marco.giordano3@graduate.univaq.it)}
\address[2]{CNIT - National Inter-university Consortium for Telecommunications (e-mail: claudia.rinaldi@univaq.it)}

\tfootnote{This work was partially supported by the EU under the Italian NRRP of NextGenerationEU, partnership on “Telecommunications of the Future” (PE00000001 - program “RESTART”).}

\begin{abstract}
Spoken Italian datasets are curated collections of audio recordings featuring native Italian speech in various contexts (e.g., spontaneous dialogues, read text, telephone conversations), often accompanied by transcriptions or linguistic annotations. They serve as foundational resources for a wide range of applications, including Automatic Speech Recognition (ASR), Text-To-Speech (TTS) synthesis, emotion detection, and broader linguistic research. Despite Italian’s status as a richly diverse Romance language—marked by significant dialectal variation—publicly available large-scale corpora have remained comparatively underrepresented when contrasted with those of major world languages such as English or Mandarin.

In this survey, we present a comprehensive examination of 66 spoken Italian datasets, highlighting their key characteristics, data collection methodologies, and annotation frameworks. We categorize the datasets by speech type (e.g., conversational, monologic, spontaneous), by source (e.g., broadcast media, telephone calls, field recordings), and by demographic or linguistic attributes (including dialects and sociolinguistic features). Our analysis uncovers critical issues around dataset scarcity, demographic underrepresentation, and restricted accessibility, limiting broader research and development efforts. To address these gaps, we propose best practices and future directions—such as expanding demographic coverage, promoting open-access models, and standardizing annotation protocols—to enrich Italian speech data resources. The complete dataset inventory is publicly available on \href{https://github.com/marco-giordano/spoken-italian-datasets}{GitHub} and archived on Zenodo, offering researchers and developers a valuable reference. By highlighting both the achievements and the shortcomings in existing resources, this work ultimately aims to foster collaboration and to spur further advancements in Italian speech technologies and linguistic research.

\end{abstract}


\titlepgskip=-15pt

\maketitle


\section{Introduction}
\subsection{Background and Motivation}
Spoken language datasets play a pivotal role in linguistic research, natural language processing (NLP) and speech technology. They provide the foundational data necessary for tasks such as automatic speech recognition (ASR), speech synthesis, and sociolinguistic analysis. Despite significant advancements in the field, linguistic resources specific to Italian remain relatively limited, particularly compared to languages such as English and Mandarin.
Italian, as a Romance language with rich phonological and dialectal diversity, offers unique challenges and opportunities for research. Its linguistic variability across regions, combined with socio-cultural nuances, makes it a compelling subject of study for developing robust speech models. The increasing demand for speech-based applications, ranging from virtual assistants to automated transcription systems, further underscores the importance of well-annotated and comprehensive Italian spoken language datasets.
Although several large-scale surveys and corpus compilations have been conducted for other languages, a focused survey detailing the variety, scope, and applications of spoken Italian datasets is notably absent, to the best of our knowledge. This paper aims to address this gap by presenting a comprehensive overview of existing resources, their methodologies, and their contributions to the field.

\subsection{Objectives of the Survey}
This survey provides a comprehensive overview of spoken Italian datasets, cataloging them by speech type, source, demographic diversity, and linguistic features. It examines methodologies for data collection, annotation, and validation, highlighting best practices and challenges. Practical applications, such as speech recognition, emotion detection, sociolinguistics, and education, are also explored.

To ensure inclusivity, the survey includes both public and commercial datasets, focusing on those actively maintained as of November 2024 with sufficient available information. While dataset quality and availability were not independently verified, the survey synthesizes information provided by creators or documentation. By identifying gaps and limitations, it offers recommendations for creating future datasets, emphasizing collaboration and standardization to expand and improve spoken Italian corpora.
While the paper highlights key datasets to illustrate the discussed categories and methodologies, the full collection comprises 66 datasets that encompass a wide range of speech types, linguistic features, and applications. This complete dataset inventory is publicly accessible via \href{https://github.com/marco-giordano/spoken-italian-datasets}{GitHub} and archived on Zenodo (DOI: 10.5281/zenodo.14246196), providing an useful and collaborative resource for researchers and developers.

\subsection{Structure of the paper}
The remainder of this paper is organized as follows.
\begin{itemize}[topsep=0pt, partopsep=0pt, itemsep=0pt, parsep=0pt]
\item Section 2 categorizes spoken Italian datasets based on speech type, source, and demographic features, providing an overview of their characteristics and applications.
\item Section 3 describes the methodologies for data collection, annotation, and transcription, highlighting best practices and common challenges.
\item Section 4 explores the practical applications of spoken Italian datasets in linguistic research, NLP, and speech technology.
\item Section 5 discusses the challenges and limitations of existing datasets, including issues related to accessibility, representation, and technical constraints.
\item Section 6 outlines future directions and recommendations to address identified gaps and improve the quality of the data set.
\item Section 7 concludes by summarizing the findings and emphasizing the importance of collaborative efforts to advance spoken Italian datasets.
\end{itemize}

\section{Categorization of Spoken Italian Datasets and Corpora}
Spoken Italian datasets vary widely in their structure, scope, and focus, reflecting the diverse needs of linguistic research and speech technology development. To provide a clear understanding of their characteristics and applications, this section categorizes the datasets based on speech type, source and context, and demographic and linguistic features. This categorization highlights the richness of available resources while identifying gaps and areas for future expansion.
\subsection{By Type of Speech}
The categorization of spoken Italian datasets by type of speech provides a structured way to understand their applications and relevance. This classification reflects the diversity of speech contexts, each serving unique research and technological objectives.
\subsubsection{Conversational Speech}
Datasets in this category capture dialogues and interactions, often in natural settings. Examples include datasets derived from interviews, informal conversations, or scripted dialogues for dialogue systems. These datasets are critical for training conversational agents, studying discourse patterns, and advancing dialogue-based NLP applications. \\ The \textbf{\textbf{ KIParla corpus}} \cite{noauthor_corpus_nodate}\cite{mauri_kiparla_nodate} provides a diverse range of conversational data collected through semi-structured interviews, table discussions, and spontaneous conversations across different cities in Italy, such as Bologna and Turin. Similarly, the \textbf{\textbf{Italian Spontaneous Dialogue Dataset}} \cite{noauthor_italian_nodate-1}, encompassing 1,797 hours of unscripted conversations recorded in domains like banking and insurance, simulates real-world interactions. Another key resource is the \textbf{\textbf{DIA - Dialogic Italian}} \cite{vietti_dia-dialogic_2021}, which focuses on spontaneous dialogues between well-acquainted speakers, recorded under controlled conditions to ensure the authenticity of natural speech patterns.

\subsubsection{Monologues}
Monologue datasets consist of continuous speech from a single speaker and are instrumental for tasks such as public speaking tools, prosody analysis, and speech synthesis \cite{radford_robust_2023},\cite{chen_wavlm_2022}. \\ The \textbf{\textbf{Italian Scripted Monologue Dataset} } \cite{noauthor_italian_nodate}, containing over 2,500 hours of pre-scripted speech recorded on mobile devices, is a prime example. Additionally, \textbf{\textbf{VoxPopuli}} \cite{noauthor_voxpopuli_2024, wang_voxpopuli_2021} provides a rich source of formal monologues, drawn from European Parliament sessions, that capture structured speech in professional settings, furthering research in areas like speech translation and recognition.

\subsubsection{Spontaneous Speech}
Spontaneous speech datasets capture the variability and unpredictability of natural, unplanned language use, making them essential for robust ASR systems and linguistic analysis \cite{dufour_spontaneous_2009}\cite{mulholland_comparison_2016}. \\ The \textbf{LABLITA corpus} \cite{cresti_lablita_2022, noauthor_lablita_nodate}, featuring recordings from face-to-face conversations and telephone calls, is a cornerstone for studying prosodic and pragmatic features in natural Italian speech. Another valuable dataset is the \textbf{\textbf{C-ORAL-ROM corpus}} \cite{noauthor_c-oral-rom_nodate}, which focuses on authentic spoken interactions in various informal contexts, including group discussions and everyday exchanges. Its rich spontaneous speech data supports research in conversational analysis and natural language processing.

\subsubsection{Read Speech}
Read speech datasets provide clean, controlled recordings, ideal for tasks like ASR training and text-to-speech systems \cite{ngueajio_hey_2022},\cite{kharitonov_speak_2023}. \\ The \textbf{\textbf{MLS dataset}} \cite{pratap_mls_2020, noauthor_mls_nodate}, for instance, features 279 hours of Italian read speech derived from public domain audiobooks. Another significant contribution comes from Mozilla’s \textbf{\textbf{Common Voice}} project \cite{noauthor_common_nodate}, a crowd-sourced effort that has collected over 1,000 hours of Italian read speech contributed by volunteers from various demographics, ensuring a diverse representation.

\subsection{By Source and Context}
The source and context of speech datasets determine their applicability and potential use cases. Categorizing datasets by source sheds light on their relevance to specific domains and scenarios aligning the dataset with particular real world scenarios. Categorizing by context impacts the dataset applicability to use cases requiring specific conditions or settings (e.g. noisy environments, emotional speech, accents etc). From an acoustic point of view, different sources (studio vs. field recordings) vary in signal-to-noise ratios, reverberation, and other acoustic properties while contexts such as background noise \cite{de_j_velasquez-martinez_combining_2023}, speaker diversity, or overlapping speech may affect pre-processing and model training strategies e.g. \cite{menne_analysis_2019}.

\subsubsection{Broadcast Media}
Broadcast media datasets capture formal and semi-formal speech styles and are particularly useful for applications in media transcription, sentiment analysis, media monitoring and sociolinguistic studies \cite{de_oliveira_identifying_2021},\cite{bell_mgb_2015}. The \textbf{\textbf{IBNC}} \cite{noauthor_ibnc_nodate}, a collection of 15 hours of transcribed Italian radio broadcasts covering diverse news topics, highlights this category. Similarly, the \textbf{Italian Lifestyle Podcast Dataset} \cite{noauthor_italian_nodate-2} includes recordings from live podcasts centered on lifestyle topics, such as mindfulness, offering conversational data in semi-formal contexts.

\subsubsection{Telephone Speech}
Telephone speech datasets are indispensable to develop telephony-based systems such as customer service automation, voice-based authentication, and call transcription tools \cite{-_exploring_2024}. The \textbf{Italian Spontaneous Dialogue Telephony Speech Dataset - Nexdata} \cite{noauthor_italianitaly_nodate} is a prominent resource, featuring 500 hours of conversational recordings made via mobile phones. These recordings include a balanced representation of regional accents and cover various familiar topics, simulating natural interactions in telephonic environments. Another notable dataset is the \textbf{Italian Spontaneous Dialogue Dataset} \cite{noauthor_italian_nodate-1} by Defined.ai, which includes telephony-based conversations recorded under both noisy and silent conditions. This dataset's emphasis on diverse environments makes it highly applicable to real-world telephony use cases. Additionally, the \textbf{Aurora Project Database} \cite{noauthor_aurora_nodate}, collected over the GSM telephone network in vehicular environments, provides speech data influenced by background noise and motion, contributing to advancements in noise-robust telephony systems. By leveraging these resources, developers can create telephony applications that maintain robust performance under diverse acoustic and contextual challenges.

\subsubsection{Social Media and Online Platforms}
Datasets sourced from social media and online platforms capture informal, dynamic, and contemporary speech styles, making them invaluable for studying spontaneous language use and training NLP systems for social media applications. The \textbf{MuST-C dataset} \cite{di_gangi_must-c_2019}, derived from TED Talks, provides a diverse collection of spoken content across multiple languages, including Italian. Its origins in freely available online speech make it a valuable resource for analyzing multilingual speech in educational and public discourse contexts. The \textbf{EMOFILM corpus} \cite{parada-cabaleiro_emofilm_2018}, for instance, extracts emotionally charged speech from Italian film dialogues, offering insights into expressive and contextually nuanced speech. The \textbf{Italian Lifestyle Podcast Dataset} \cite{noauthor_italian_nodate-2}, curated from live Italian podcasts, provides conversational speech data focusing on lifestyle topics, such as mindfulness and well-being. This dataset reflects authentic interactions in semi-formal contexts, making it particularly useful for training models for sentiment analysis and topic classification. Another resource is the \textbf{Common Voice} project \cite{noauthor_common_nodate} by Mozilla, which, although primarily a read-speech corpus, incorporates recordings made by diverse contributors using web-based tools. This dataset often overlaps with the informality seen in user-generated content online. These datasets collectively support applications ranging from automated content moderation to training voice assistants that can understand diverse speech patterns found on online platforms.

\subsubsection{Field Recordings}
Field recordings capture natural speech in authentic, often uncontrolled environments, making them essential for studying real-world linguistic diversity and for applications requiring robust ASR systems. The \textbf{VIVALDI corpus} \cite{noauthor_vivaldi_nodate}, for instance, focuses on regional dialects and minority languages, offering rich insights into Italy’s linguistic diversity through recordings gathered from public and informal settings. Similarly, the \textbf{DEMoS} \cite{parada-cabaleiro_demos_2020, parada-cabaleiro_demos_2019} corpus includes recordings that, although conducted in controlled settings, elicit emotions that mimic natural interactions. This approach bridges the gap between spontaneous and emotionally charged speech, adding valuable dimensions to field-recording-like data. Currently, no additional datasets in this survey explicitly feature field recordings as a primary characteristic, underscoring a potential gap in the representation of naturalistic and spontaneous speech in public or outdoor environments within the scope of Italian spoken corpora.

\subsection{By Demographic and Linguistic Features}
Spoken Italian datasets can differ greatly in both demographic and linguistic coverage. When a dataset skews toward a specific demographic—such as a particular gender, age group, or ethnicity—it risks producing biased models. Conversely, ensuring balanced demographic representation helps models perform equitably across all groups, thereby preventing unfair outcomes. However, if the goal is to study a specific region or ethnic community, it is important to note that speech patterns can vary in rhythm and intonation even within the same language. Properly characterizing these features is essential for creating a dataset that serves the needs of targeted applications within that community. Also, the lack of linguistic diversity in datasets leads to models that fail in real-world settings where varied accents or code-switching (mixing languages) are common. These distinctions highlight the sociolinguistic dimensions of speech data.

\subsubsection{Regional Varieties and Dialects}
Datasets highlighting regional varieties and dialects are essential to preserve linguistic diversity and understand the intricate patterns of variation within Italian. These datasets enable researchers to analyze phonological, syntactic, and lexical differences across regions and promote the development of language technologies that cater to regional linguistic characteristics. The \textbf{VIVALDI corpus} \cite{noauthor_vivaldi_nodate} is a cornerstone resource in this domain, offering extensive coverage of Italian dialects and minority languages such as Lombard, Sicilian, and Sardinian. By documenting speech from over 14 regions, this dataset supports studies in dialectology and sociolinguistics, as well as efforts in language preservation. The\textbf{ KIParla corpus} \cite{mauri_kiparla_nodate, noauthor_corpus_nodate}, while focused on standard Italian, includes contributions from speakers across various regions, reflecting a wide spectrum of regional accents and idiomatic expressions. This diversity makes it a valuable resource for analyzing how regional varieties influence spoken language in different contexts. Another noteworthy dataset is the \textbf{LABLITA corpus} \cite{cresti_lablita_2022, noauthor_lablita_nodate}, which, in addition to its focus on spontaneous and natural speech, incorporates recordings from different Italian regions. This regional emphasis allows researchers to explore how standard Italian interacts with regional speech patterns in everyday communication. The \textbf{DEMoS} corpus \cite{parada-cabaleiro_demos_2020, parada-cabaleiro_demos_2019}, while primarily centered on emotional speech, also includes instances of regional accents among its participants, providing an additional dimension of linguistic diversity. Similarly, the \textbf{Common Voice} project \cite{noauthor_common_nodate} features contributions from volunteers representing various regions of Italy, ensuring a broad geographic coverage that reflects the nation's linguistic heterogeneity. Finally, the \textbf{C-ORAL-ROM corpus} \cite{noauthor_c-oral-rom_nodate}, though multilingual, includes a substantial component of Italian speech from different regional contexts, enabling comparative studies of dialectal usage within the broader Romance language family.

\subsubsection{Sociolinguistic Variation}
Sociolinguistic datasets capture variations in speech influenced by factors such as age, gender, socio-economic status, and educational background. These datasets are critical for understanding how social variables shape linguistic practices and for building inclusive language technologies. The\textbf{ KIParla corpus} \cite{mauri_kiparla_nodate, noauthor_corpus_nodate} is a key resource for sociolinguistic research, featuring participants from diverse socio-economic, educational, and regional backgrounds. By including speakers of different ages and genders, this dataset allows for nuanced analyses of how social factors interact with linguistic features in Italian. The \textbf{LABLITA corpus} \cite{cresti_lablita_2022, noauthor_lablita_nodate} further enriches sociolinguistic research with its recordings of natural speech collected from various social contexts, such as face-to-face conversations and telephone calls. The dataset includes participants across a wide age range and from different social strata, offering a comprehensive view of spoken Italian as influenced by sociolinguistic diversity. The \textbf{DEMoS corpus} \cite{parada-cabaleiro_demos_2020, parada-cabaleiro_demos_2019}, focused on emotional speech, also contributes to sociolinguistic studies by involving participants aged 20 to 64 from vaious socio-economic backgrounds. This range supports the examination of how emotion and sociolinguistic variables interact in speech. Another significant dataset is the \textbf{ITALIC corpus} \cite{koudounas_italic_2023,koudounas_italic_2023-1}, which features crowd-sourced recordings from 70 speakers spanning different Italian regions, age groups, and genders. The diversity of contributors ensures broad coverage of sociolinguistic variation, making it suitable for research in intent classification and linguistic variation. Lastly, the \textbf{Common Voice} project \cite{noauthor_common_nodate} by Mozilla includes contributions from a wide demographic spectrum, with metadata detailing the age, gender, and accents of speakers. This sociolinguistic richness makes the dataset valuable for studying how demographic factors influence speech patterns in Italian.

\subsubsection{Bilingual and Multilingual Corpora}
Bilingual and multilingual corpora enable cross-linguistic studies and support the development of multilingual speech technologies, such as machine translation, speech synthesis, and multilingual ASR systems. These datasets are particularly valuable for studying language contact, code-switching, and comparative linguistics. The \textbf{Europarl-ST corpus} \cite{noauthor_europarl_nodate, iranzo-sanchez_europarl-st_2020}, derived from European Parliament debates, includes approximately 64.18 hours of Italian speech alongside translations into eight other European languages. This resource is instrumental for speech translation research, providing aligned data for tasks as machine learning and language modeling. The \textbf{PortMedia French and Italian Corpus} \cite{noauthor_portmedia_nodate}, focusing on tourist information and reservation dialogues, provides bilingual data that is highly relevant for domain-specific studies. With 604 Italian dialogues aligned with French data, this corpus supports applications in multilingual dialogue systems and domain-specific NLP. The \textbf{MuST-C corpus} \cite{di_gangi_must-c_2019}, a multilingual speech-to-text dataset, provides high-quality audio-text pairs for automatic speech translation, including an Italian subset. This dataset is instrumental in advancing research on cross-linguistic speech technologies and translation systems. The \textbf{M-AILABS Speech Dataset} \cite{celeste_witchzard_m-ailabs-dataset_2024} also includes an Italian subset as part of its multilingual collection. Derived from public domain audiobooks, the dataset spans multiple languages, enabling comparative studies in read speech and providing training data for multilingual speech systems. The \textbf{MULTEXT Prosodic Database} \cite{noauthor_multext_nodate} offers multilingual resources, including Italian, French, German, English, and Spanish. Its focus on prosodic annotations makes it particularly useful for studying intonation patterns across languages, supporting research in paralinguistics and language comparison. Lastly, the \textbf{C-ORAL-ROM corpus} \cite{noauthor_c-oral-rom_nodate} integrates Italian alongside other Romance languages, such as French, Spanish, and Portuguese. Its recordings of spontaneous speech in diverse settings allow for comparative studies of spoken language dynamics within the Romance language family.

\section{Data Collection and Annotation Methodologies}
\subsection{Data Collection Techniques}
Robust spoken language datasets depend on effective data collection strategies, which typically involve selecting representative speaker samples, designing appropriate recording protocols, and ensuring high-quality audio capture. The methods used can vary widely based on the dataset purpose, ranging from controlled studio environments to spontaneous field recordings. Proper planning in data collection ensures the resulting datasets meet the specific requirements for linguistic research, ASR training, or speech technology applications.
\subsubsection{Recording Methods}
Recording methods play a crucial role in determining the quality and applicability of speech datasets. They can range from controlled environments with professional equipment to real-world settings with ambient noise, each providing unique advantages depending on the research goals. The \textbf{Apasci dataset} \cite{noauthor_apasci_nodate} exemplifies a highly controlled approach, with recordings conducted in an insulated room using a Sennheiser MKH 416 T microphone. This setup ensures high-fidelity audio, making the dataset suitable for phonetic research and speech recognition development. In contrast, the \textbf{Aurora Project Database} \cite{noauthor_aurora_nodate} focuses on real-world scenarios by capturing speech in vehicles under various driving conditions. Speech was recorded using multiple microphones connected to computers and mobile phones, providing data valuable for noise-robust speech recognition and in-car voice command systems. The \textbf{Italian Kids Speech Recognition Corpus} (Desktop) \cite{noauthor_italian_nodate-12} features recordings of children in a quiet office setting, using high-quality microphones to ensure clarity while focusing on a demographic often underrepresented in speech corpora. The \textbf{Italian Spontaneous Dialogue Dataset} \cite{noauthor_italian_nodate-1} takes a different approach by recording spontaneous conversations via telephony in both noisy and silent environments. This diversity in conditions supports the development of telephony-based ASR systems capable of handling real-world challenges. Another example, the \textbf{MULTEXT Prosodic Database} \cite{noauthor_multext_nodate}, incorporates specialized methods by recording passages using standardized protocols to focus on prosody. The careful design of the recording environment ensures that the dataset captures nuanced prosodic features essential for speech analysis. The \textbf{M-AILABS Speech Dataset} \cite{celeste_witchzard_m-ailabs-dataset_2024}, sourced from LibriVox audiobooks, utilizes volunteer-based recordings. While the setup varies in quality due to its crowdsourced nature, it provides extensive coverage of read speech, contributing significantly to multilingual ASR training.

\subsubsection{Participant Selection}
Participant selection is a critical step in data collection, ensuring that speech datasets are representative of the target population and suitable for their intended applications. Factors such as age, gender, regional origin, socio-economic background, and linguistic proficiency are carefully considered to achieve demographic diversity and coverage. The\textbf{ KIParla corpus} \cite{mauri_kiparla_nodate, noauthor_corpus_nodate} exemplifies a well-rounded approach to participant selection, drawing speakers from diverse age groups, genders, educational levels, and regional origins. This ensures the dataset captures sociolinguistic variations across Italy, making it a valuable resource for comprehensive linguistic analysis. The \textbf{DEMoS corpus} \cite{parada-cabaleiro_demos_2020, parada-cabaleiro_demos_2019}, focused on emotional speech, employs a more targeted selection strategy. It includes 68 native Italian speakers aged 20 to 64, chosen specifically for their vocal expressivity. This careful selection enhances the dataset’s utility for research in speech emotion recognition and affective computing. For datasets targeting specific demographics, the \textbf{Italian Kids Speech Recognition Corpus} \cite{noauthor_italian_nodate-12} is a prime example. It exclusively features recordings from children, ensuring its relevance for applications such as educational tools and child-specific speech recognition systems. The \textbf{M-AILABS Speech Dataset} \cite{celeste_witchzard_m-ailabs-dataset_2024} adopts a crowdsourced model, relying on volunteer contributors from diverse backgrounds. While this approach introduces variability in speaker demographics and recording conditions, it also ensures a wide-ranging representation, particularly for read speech. Similarly, the \textbf{Common Voice} project \cite{noauthor_common_nodate} by Mozilla encourages open participation, with contributors from various age groups, genders, and regional accents. The associated metadata provides valuable information about the speakers, enabling detailed demographic analysis and customization of speech models. In the case of dialectal and minority language preservation, the \textbf{VIVALDI corpus} \cite{noauthor_vivaldi_nodate} focuses on native speakers of regional dialects and minority languages across Italy. By selecting participants based on linguistic heritage, the dataset provides authentic representations of Italy’s linguistic diversity.

\subsection{Annotation and Transcription Processes}
Annotation and transcription add significant value to raw speech data, enabling downstream applications like speech recognition, phonetic analysis, and linguistic studies.
\subsubsection{Transcription Standards}
Transcription standards are critical for ensuring that speech data can be accurately analyzed and utilized across diverse applications. Depending on the dataset's purpose, transcription may be orthographic \cite{alfano_volip_nodate}, phonetic \cite{savino_methods_nodate}, or include additional linguistic annotations such as prosody or disfluencies. Proper transcription practices ensure data consistency, facilitate downstream processing, and enable interoperability between datasets. The debate between orthographic and phonetic transcription continues to be a focus of research, with studies such as Bryła-Cruz's \cite{bryla-cruz_more_2022} emphasizing the limitations of orthographic systems for representing phonological nuances. Phonetic transcriptions, particularly those employing the International Phonetic Alphabet (IPA), are often preferred for their precision, though they require specialized expertise.
The \textbf{VIVALDI corpus} \cite{noauthor_vivaldi_nodate} incorporates a combination of orthographic and phonetic transcription to document dialectal variations and minority languages, providing researchers with detailed linguistic insights. The \textbf{LABLITA corpus} \cite{cresti_lablita_2022, noauthor_lablita_nodate} offers a more complex annotation approach, combining orthographic transcription with prosodic features such as intonation patterns and stress. This multi-layered annotation supports studies in discourse analysis and prosodic modeling, particularly for spontaneous speech. Another dataset, \textbf{EMOFILM} \cite{parada-cabaleiro_emofilm_2018}, employs orthographic transcription for its emotional speech data, aligning text with expressed emotions for sentiment and emotion recognition research. By capturing nuanced speech elements, this dataset bridges the gap between linguistic and emotional analysis. The importance of phonetic transcription for capturing pronunciation details is highlighted in \textbf{M-AILABS Speech Dataset} \cite{celeste_witchzard_m-ailabs-dataset_2024}, which provides phonetic alignments for multilingual audiobooks, including Italian. These transcriptions facilitate tasks requiring precise phonological information, such as text-to-speech synthesis.

\subsubsection{Annotation Levels}
Annotation levels refer to the granularity and types of linguistic information included in speech datasets. These levels can range from basic lexical transcriptions to detailed syntactic, semantic, and prosodic annotations, depending on the dataset's intended use. Comprehensive annotation enriches the dataset, making it more versatile for applications in linguistics, speech technology, and natural language processing. Automated annotation tools, such as those discussed in Stan's study \cite{stan_recoapy_2020}, have increasingly supported the generation of multi-level annotations. These tools streamline the annotation process, particularly for large-scale multilingual datasets. The \textbf{LABLITA corpus} \cite{cresti_lablita_2022, noauthor_lablita_nodate} exemplifies multi-level annotation, combining orthographic transcription with prosodic and pragmatic features. This dataset includes annotations for intonation, stress patterns, and discourse markers, enabling research in prosody and discourse analysis for spontaneous speech. Its multi-layered approach provides a rich resource for examining the interaction of linguistic and prosodic features. Similarly, the \textbf{VIVALDI corpus} \cite{noauthor_vivaldi_nodate} incorporates lexical, syntactic, and phonetic annotations, with additional layers for dialectal and regional features. By documenting speech from various Italian dialects, the dataset supports studies in sociolinguistics and dialectology, while its syntactic annotations facilitate syntax-based linguistic research. In the\textbf{ KIParla corpus} \cite{mauri_kiparla_nodate, noauthor_corpus_nodate}, annotations include lexical content, turn-taking cues, and speaker metadata such as gender and age. This multi-level annotation enables the study of conversational dynamics and sociolinguistic variation, making it a valuable resource for analyzing spoken Italian in social contexts. The \textbf{Italian Kids Speech Recognition Corpus} \cite{noauthor_italian_nodate-12} employs lexical and phonetic annotation to support child-specific ASR development. This dual-layer annotation is particularly important for modeling the phonological characteristics of children's speech, which differ from adult speech in pronunciation, prosody, and syntax. Advanced datasets like \textbf{MULTEXT Prosodic Database} \cite{noauthor_multext_nodate} integrate prosodic, syntactic, and semantic annotations. This database focuses on cross-linguistic prosody research, offering detailed annotations for stress patterns, pitch contours, and syntactic structures. Such comprehensive annotation facilitates comparative studies across languages and applications in multilingual ASR systems. For emotional and expressive speech, the \textbf{DEMoS corpus} \cite{parada-cabaleiro_demos_2020, parada-cabaleiro_demos_2019} includes annotations for emotion categories, intensity levels, and speaker affect. These annotation levels enhance its utility in affective computing and emotion recognition technologies.

\subsubsection{Tools and Software Used}
The quality and efficiency of speech data annotation depend significantly on the tools and software employed. These tools facilitate tasks such as transcription, segmentation, and annotation of linguistic and prosodic features. Below are key tools and software used in annotating spoken Italian datasets, along with their contributions and applications. One of the most widely used tools is Praat \cite{noauthor_praat_nodate}, a software designed for phonetic analysis and annotation. Praat enables the annotation of pitch, intensity, and formant structures, making it indispensable for datasets like the \textbf{LABLITA corpus} \cite{cresti_lablita_2022, noauthor_lablita_nodate}, where prosodic features such as intonation patterns and stress need to be analyzed in detail.
ELAN \cite{noauthor_elan_nodate}, developed by the Max Planck Institute for Psycholinguistics, is another key tool. ELAN supports multi-tier annotations, enabling researchers to align audio with lexical, syntactic, and semantic layers. It is frequently used in datasets like the \textbf{VIVALDI corpus} \cite{noauthor_vivaldi_nodate}, where dialectal and regional features require detailed annotation across multiple linguistic dimensions.
The WebMAUS tool \cite{noauthor_webmaus_nodate}, developed by the Bavarian Archive for Speech Signals, automates phonetic segmentation and alignment. Its ability to handle multilingual data, including Italian, makes it a valuable resource for datasets such as the\textbf{ KIParla corpus} \cite{mauri_kiparla_nodate, noauthor_corpus_nodate}, where phonetic alignment is critical for analyzing conversational dynamics.
For corpora focusing on emotion and expressivity, tools like Emu-SDMS \cite{winkelmann_emu-sdms_2017} provide integrated solutions for speech database management and annotation. The \textbf{DEMoS corpus} \cite{parada-cabaleiro_demos_2020, parada-cabaleiro_demos_2019} benefits from Emu-SDMS's ability to handle emotional speech annotations, including intensity levels and affective markers.
SPPAS - SPeech Phonetization Alignment and Syllabification \cite{noauthor_sppas_nodate, bigi_sppas_2012} is a specialized tool for generating annotations for speech synthesis applications. It has been applied in datasets such as the \textbf{Italian Scripted Monologue Dataset}  \cite{noauthor_italian_nodate}, where precise lexical and phonetic alignment is crucial for synthesizing high-quality speech.
Finally, Annotald \cite{noauthor_annotald_nodate}, a tool developed for syntactic and morphological annotation, has been utilized in datasets requiring deep linguistic analysis, such as the \textbf{MULTEXT Prosodic Database} \cite{noauthor_multext_nodate}.

\subsection{Quality Control and Validation}
Quality control and validation are essential steps in the creation of spoken language datasets, ensuring that the data is accurate, consistent, and reliable for research and technology development. Techniques for maintaining quality vary depending on the dataset’s purpose, size, and complexity but typically include manual verification, automated checks, and statistical analyses.

For phonetic and lexical transcriptions, grapheme-to-phoneme (G2P) converters are often used to verify pronunciation patterns against standardized rules. In projects employing tools like RECOApy \cite{stan_recoapy_2020}, automated G2P conversion identifies discrepancies in phonetic transcriptions, flagging them for manual review, but according to our current knowledge, none of the documented datasets made use of this tool in validation processes.

One common approach is manual validation, where trained linguists or annotators review a subset of the dataset for errors or inconsistencies. For example, in the\textbf{ KIParla corpus} \cite{mauri_kiparla_nodate, noauthor_corpus_nodate}, annotations were manually cross-checked by multiple experts to ensure the accuracy of transcriptions and metadata, particularly for conversational dynamics and turn-taking cues.

Inter-annotator agreement (IAA) \cite{artstein_inter-annotator_2017} is another widely used metric for evaluating annotation consistency. In the \textbf{LABLITA corpus} \cite{cresti_lablita_2022, noauthor_lablita_nodate}, IAA scores were computed to ensure that annotations for prosodic and discourse features met a high standard of agreement among annotators. High IAA scores indicate that the annotation scheme is robust and reproducible across different annotators.

Automated validation tools play a crucial role in handling large datasets. Tools like, already cited, WebMAUS \cite{noauthor_webmaus_nodate} and Praat \cite{noauthor_praat_nodate} can automatically align phonetic transcriptions with audio files and detect mismatches or misalignments. For datasets like the \textbf{VIVALDI corpus} \cite{noauthor_vivaldi_nodate}, automated checks ensure that dialectal transcriptions align correctly with the speech recordings.

Noise and artifact detection is crucial for datasets collected in uncontrolled environments, such as the \textbf{Italian Spontaneous Dialogue Dataset} \cite{noauthor_italian_nodate-1}. Techniques such as spectral analysis and noise filtering are applied to identify and minimize the impact of background noise and recording artifacts. For example, Audacity \cite{noauthor_audacity_nodate} and other audio editing tools are used to clean recordings before annotation.

Sampling and statistical validation ensure the dataset is representative of the target population. In the \textbf{Common Voice} project \cite{noauthor_common_nodate}, speaker demographics are periodically analyzed to confirm that the dataset includes sufficient diversity in accents, ages, and genders. Statistical checks also ensure even representation of various phonetic and lexical features.

Cross-validation is employed in datasets used for machine learning tasks. For instance, the \textbf{M-AILABS Speech Dataset} \cite{celeste_witchzard_m-ailabs-dataset_2024} uses cross-validation during model training to evaluate the dataset’s reliability for ASR systems. This ensures that the dataset does not contain biases or inconsistencies that could compromise model performance.

Lastly, version control and iterative improvements are increasingly common in dataset projects. Open-source datasets, such as \textbf{Common Voice} \cite{noauthor_common_nodate}, allow contributors to submit corrections or improvements, ensuring the dataset evolves and maintains quality over time.

In Table \ref{tab:dataset_summary}, we summarize the characteristics of the documented datasets w.r.t. size, source and context, type of speech, demographic and linguistic features and transcription standard.

\begin{table*}
\centering
\caption{Summary of Selected Datasets}
\label{tab:dataset_summary}
\resizebox{\linewidth}{!}{%
\begin{tabularx}{\textwidth}{|X|X|X|X|X|X|}
\hline
\textbf{Dataset Name} & \textbf{Size} & \textbf{Source and Context} & \textbf{Type of Speech} & \textbf{Demographic \& Linguistic Features} & \textbf{Transcription Standard} \\
\hline
\textbf{Apasci} \cite{noauthor_apasci_nodate} & 641 minutes & Phonetically rich sentences and digits & Phonetic speech & Standard Italian, no sociolinguistic or multilingual focus & Not specified \\
\hline
\textbf{AURORA Project database} \cite{noauthor_aurora_nodate} & 2,200 utterances & In-car recordings & Connected digits & Standard Italian, no specific sociolinguistic features & Not specified \\
\hline
\textbf{C-ORAL-ROM} \cite{noauthor_c-oral-rom_nodate} & 1,200,000 words & Spontaneous speech in Romance languages & Spontaneous & European variants of Romance languages & Orthographic \\
\hline
\textbf{Common Voice} \cite{noauthor_common_nodate}& 1,000 hours & Crowdsourced contributions & Read speech & Various regional accents & Orthographic \\
\hline
\textbf{DEMoS - an Italian emotional speech corpus} \cite{parada-cabaleiro_demos_2019}\cite{parada-cabaleiro_demos_2020}& 9,697 recordings & Mood induction procedures & Emotional speech & Native Italian speakers & Orthographic \\
\hline
\textbf{DIA - Dialogic Italian} \cite{vietti_dia-dialogic_2021} & 10 hours & Dialogues between acquainted speakers & Dialogic & Speakers from Bolzano area & Orthographic \\
\hline
\textbf{EMOFILM} \cite{parada-cabaleiro_emofilm_2018} & Details not specified & Film dialogues & Emotional & Regional and multilingual & Orthographic \\
\hline
\textbf{Europarl-ST} \cite{noauthor_europarl_nodate}\cite{iranzo-sanchez_europarl-st_2020}& 64.18 hours & Parliamentary debates & Formal speeches & European Parliament speakers & Orthographic \\
\hline
\textbf{IBNC - An Italian Broadcast News Corpus} \cite{noauthor_ibnc_nodate} & 15 hours & Broadcast news & Read and spontaneous & Standard Italian & Orthographic \\
\hline
\textbf{Italian Spontaneous Dialogue Telephony Speech Dataset} \cite{noauthor_italianitaly_nodate} & 500 hours & Telephone dialogues & Conversational & Regional accents, balanced gender & Orthographic \\
\hline
\textbf{Italian Kids Speech Recognition Corpus (Desktop)} \cite{noauthor_italian_nodate-12} & 4.9 hours & Children's speech & Children's speech & Children only & Orthographic \\
\hline
\textbf{Italian Lifestyle Podcast Dataset} \cite{noauthor_italian_nodate-2} & 60 hours & Podcasts & Spontaneous & Various demographics & Not specified \\
\hline
\textbf{Italian Scripted Monologue Dataset} \cite{noauthor_italian_nodate} & 2,531 hours & Scripted monologues & Scripted monologues & Standard Italian, diverse speakers & Orthographic \\
\hline
\textbf{Italian Spontaneous Dialogue Dataset} \cite{noauthor_italian_nodate-1} & 1,797 hours & Dialogues & Dialogues & Regional accents & Orthographic \\
\hline
\textbf{KIParla - Corpus of Spoken Italian} \cite{noauthor_corpus_nodate, mauri_kiparla_nodate} & 150+ hours & Conversational contexts & Conversational & Regional varieties, diverse sociolinguistics & Simplified Jefferson system \\
\hline
\textbf{LABLITA Corpus} \cite{noauthor_lablita_nodate, cresti_lablita_2022} & 400,000 words & Various communicative situations & Spontaneous & Regional varieties, socio-economic variations & L-AcT framework \\
\hline
\textbf{M-AILABS Speech Dataset} \cite{celeste_witchzard_m-ailabs-dataset_2024} & 108 hours & Public domain audiobooks & Read speech & Standard Italian & Orthographic \\
\hline
\textbf{MLS - Multilingual Large Scale Dataset} \cite{noauthor_mls_nodate} & 279.43 hours & LibriVox audiobooks & Read speech & Primarily standard Italian & Orthographic \\
\hline
\textbf{MULTEXT Prosodic Database} \cite{noauthor_multext_nodate} & 4+ hours & EUROM.1 corpus & Read speech & Multilingual European languages & Orthographic \\
\hline
\textbf{MuST-C} \cite{di_gangi_must-c_2019}& 385 hours & TED talks & TED talks & Italian TED talks & Orthographic \\
\hline
\textbf{PortMedia French and Italian Corpus} \cite{noauthor_portmedia_nodate}& Details not specified & French and Italian dialogues & Dialogues & Standard Italian & Orthographic \\
\hline
\textbf{VIVALDI - Vivaio delle lingue e dei dialetti d'Italia} \cite{noauthor_vivaldi_nodate}& Data from 14 regions & Dialects across Italian regions & Spontaneous and elicited & Italian dialects & Phonetic \\
\hline
\textbf{VoxPopuli} \cite{noauthor_voxpopuli_2024, wang_voxpopuli_2021} & 400,000 hours & Parliamentary event recordings & Formal & Standard European languages & Orthographic \\
\hline
\end{tabularx}
} 

\end{table*}

\section{Applications of Spoken Italian Datasets}
Spoken Italian datasets could serve as foundational resources across a range of disciplines, enabling advancements in both theoretical research and practical applications. Their diverse scope, encompassing regional varieties, demographic diversity, and speech types, allows for broad exploration in linguistics, speech technologies, education, and cultural studies. This section outlines key areas where these datasets may have a significant impact. In table \ref{tab:applications} we summarize the relevance of 23 datasets w.r.t. the application fields outlined in this section.
\subsection{Linguistic Research}
Spoken Italian datasets are pivotal in linguistic research by providing empirical data for the study of phonetics, phonology, syntax, and sociolinguistics. Their rich annotations and diverse content enable researchers to uncover patterns and variations in spoken language use.

\subsubsection{Phonetics and Phonology}
Datasets such as the \textbf{LABLITA corpus} \cite{cresti_lablita_2022, noauthor_lablita_nodate}, EMOVO corpus \cite{costantini_emovo_nodate}, and \textbf{VIVALDI corpus} \cite{noauthor_vivaldi_nodate} are instrumental in studying Italian intonation, stress patterns \cite{pucci_speech_2023}, and prosody \cite{federico_evaluating_2020}. These resources facilitate detailed analyses of regional and dialectal phonological differences, shedding light on the dynamics of spoken Italian in various communicative settings.

\subsubsection{Syntax and Morphology}
Annotated datasets like \textbf{KIParla} \cite{mauri_kiparla_nodate, noauthor_corpus_nodate} provide valuable insights into spoken syntax, including ellipsis, word order variations, and interactional markers typical of conversational Italian. The integration of morphosyntactic annotations allows for the exploration of spontaneous speech structures, contributing to theoretical advancements in syntax.

\subsubsection{Sociolinguistics}
Sociolinguistic studies can greatly benefit from datasets such as DEMoS \cite{parada-cabaleiro_demos_2020, parada-cabaleiro_demos_2019} and KIParla \cite{mauri_kiparla_nodate, noauthor_corpus_nodate}, which incorporate metadata on speakers’ age, gender, and regional origins. These datasets enable investigations into how social variables influence linguistic behavior, such as the use of dialectal features, gendered speech patterns, and age-related linguistic variation.

\subsubsection{Dialectology}
Resources like the \textbf{VIVALDI corpus} \cite{noauthor_vivaldi_nodate}, focusing on regional varieties and minority languages, have revitalized research on Italy’s dialectal diversity. These datasets document phonological and lexical variations across dialects, contributing to efforts to preserve endangered linguistic traditions \cite{boula_de_mareuil_for_2021} \cite{huszthy_untamed_2017} \cite{la_quatra_speech_2024}.

\subsubsection{Discourse and Pragmatics}
With their rich conversational content, datasets like the \textbf{Italian Spontaneous Dialogue Dataset} \cite{noauthor_italian_nodate-1} support the study of pragmatic phenomena such as turn-taking, interruptions, and politeness strategies. This research has implications for both discourse theory and the development of natural conversation models.

\subsection{Natural Language Processing (NLP) and Speech Technologies}
Spoken Italian datasets play a critical role in advancing Natural Language Processing (NLP) and speech technologies, providing the foundation for building and refining various speech-based systems. These applications span automatic speech recognition (ASR), speech synthesis, machine translation, and sentiment analysis.

\subsubsection{Automatic Speech Recognition (ASR)}
ASR systems rely heavily on high-quality annotated speech datasets for training and testing. Resources like the \textbf{Italian Spontaneous Dialogue Dataset} \cite{noauthor_italian_nodate-1} and \textbf{Common Voice} project \cite{noauthor_common_nodate} offer diverse and realistic speech samples, enabling ASR models to handle spontaneous speech, regional accents, and environmental noise. These datasets may support the development of voice assistants, automated transcription services, and accessibility tools for Italian speakers like in \cite{kamble_custom_2023}, where \textbf{Common Voice} has been applied to Hindi.

\subsubsection{Speech Synthesis}
Text-to-speech (TTS) systems use read-speech datasets to model the nuances of human pronunciation and intonation. The \textbf{Italian Scripted Monologue Dataset}  \cite{noauthor_italian_nodate}, \textbf{Common Voice} dataset \cite{noauthor_common_nodate}, and \textbf{MLS dataset} \cite{pratap_mls_2020, noauthor_mls_nodate} may be widely used for this purpose, as their controlled recordings ensure high fidelity in synthesized speech. These applications may find use in audiobooks, virtual agents, and assistive technologies for individuals with disabilities \cite{favaro_itacotron_nodate}.

\subsubsection{Machine Translation}
Multilingual datasets like \textbf{VoxPopuli} \cite{noauthor_voxpopuli_2024, wang_voxpopuli_2021} and \textbf{C-ORAL-ROM} \cite{noauthor_c-oral-rom_nodate} provide aligned Italian speech and text data for training translation systems. The \textbf{MuST-C dataset} \cite{di_gangi_must-c_2019} plays a critical role in automatic speech translation (AST) research. Its Italian subset offers aligned audio and text for tasks such as ASR and machine translation, bridging the gap between spoken language and multilingual NLP applications. These corpora enable advancements in spoken language translation, particularly for tasks requiring real-time performance, such as multilingual meetings or subtitling.

\subsubsection{Sentiment and Emotion Analysis}
Datasets such as \textbf{DEMoS} \cite{parada-cabaleiro_demos_2020, parada-cabaleiro_demos_2019} and \textbf{EMOFILM corpus} \cite{parada-cabaleiro_emofilm_2018} may support research in recognizing emotions and sentiment from speech. These resources provide labeled data on affective states, enabling the development of sentiment analysis tools for applications in customer service, mental health, and marketing.

\subsubsection{Speech-Based Personalization}
With metadata capturing speaker demographics, datasets like KIParla \cite{mauri_kiparla_nodate, noauthor_corpus_nodate} may facilitate the development of systems that adapt to individual users, such as accent-specific voice assistants or tailored language learning applications.

\subsubsection{Robustness in Noisy Environments}
Resources like the Italian \textbf{SpeechDat-Car} database \cite{noauthor_italian_nodate-7} may contribute to building noise-robust speech technologies. These datasets include recordings in challenging environments, such as moving vehicles, which are critical for improving ASR and TTS systems in real-world conditions.

\subsection{Educational and Cultural Studies}
Spoken Italian datasets serve as valuable resources for educational and cultural initiatives, fostering linguistic understanding and promoting the preservation of Italy’s rich linguistic heritage. Their applications extend to language learning, dialect preservation, and cultural studies.

\subsubsection{Language Learning Tools}
Datasets like \textbf{Common Voice} \cite{noauthor_common_nodate} and \textbf{KIParla} \cite{mauri_kiparla_nodate, noauthor_corpus_nodate} may be used in developing tools for Italian language learners. These datasets provide authentic examples of spoken Italian, including regional accents and conversational styles, which enhance learners’ comprehension and pronunciation skills. For instance, interactive language learning applications and voice-controlled language tutors may leverage these datasets to create personalized, engaging experiences.

\subsubsection{Children’s Language Development}
The \textbf{Italian Kids Speech Recognition Corpus} \cite{noauthor_italian_nodate-12} plays a significant role in studying language acquisition among children. By capturing age-specific speech patterns, this dataset supports research into the linguistic development of young speakers and informs the design of child-friendly educational tools, such as reading apps and interactive games.

\subsubsection{Cultural Analysis and Documentation}
Datasets like the \textbf{EMOFILM corpus} \cite{parada-cabaleiro_emofilm_2018}, which captures emotionally charged speech from Italian films, offer unique insights into cultural expressions and emotional communication. These resources enable studies in film and media analysis, helping researchers understand how Italian culture is reflected in its language and storytelling.

\subsubsection{Applications in Linguistic and Cultural Heritage Projects}
Spoken corpora are integral to projects that aim to celebrate and maintain Italy’s linguistic and cultural heritage. For example, museums and cultural institutions may use speech synthesis models trained or fine-tuned on specific audio datasets to create interactive exhibits that showcase regional accents, historical speech patterns, and oral traditions.

\begin{landscape}
\begin{table}[ht]

\centering
\caption{Datasets vs.\ Applications (Section IV). A check mark (\checkmark) indicates relevance to that specific sub-subsection. The table refers to a subset (23 over 66) of reported datasets. The complete datasets collection can be found on \href{https://github.com/marco-giordano/spoken-italian-datasets}{GitHub}.}
\label{tab:applications}
\resizebox{\linewidth}{!}{%
\begin{tabular}{|p{2.5cm}|c|c|c|c|c|c|c|c|c|c|c|c|c|c|c|}
\hline
& \multicolumn{5}{|c|}{\textbf{4.1 Linguistic Research}} 
& \multicolumn{6}{|c|}{\textbf{4.2 NLP \& Speech Tech}} 
& \multicolumn{4}{|c|}{\textbf{4.3 Educational \& Cultural}}\\ \hline
\textbf{Dataset} 
& \parbox{1.2cm}{Phonetics Phonology} 
& \parbox{1cm}{Syntax Morphology} 
& \parbox{1cm}{Sociol.} 
& \parbox{1cm}{Dialect.} 
& \parbox{1.2cm}{Discourse Pragmatics}
& \parbox{1cm}{ASR} 
& \parbox{1cm}{TTS} 
& \parbox{1.2cm}{Machine Translation} 
& \parbox{1cm}{Sentim. Emotion} 
& \parbox{1cm}{Person.} 
& \parbox{1.2cm}{Noise Robustness}
& \parbox{1.4cm}{Language Learning Tools} 
& \parbox{1cm}{Child.'s Language Dev.} 
& \parbox{1cm}{Cultural Analysis} 
& \parbox{1cm}{Heritage Projects}
\\ 
\hline

\textbf{Apasci} \cite{noauthor_apasci_nodate} 
 & \checkmark &        &        &        &      
 & \checkmark &        &        &        &       
 &        &        &        &        &  
 \\
 \hline

\textbf{Aurora Project} \cite{noauthor_aurora_nodate} 
 &        &        &        &        &       
 & \checkmark &        &        &        &       
 & \checkmark &        &        &        &  
 \\
 \hline
 
\textbf{C-ORAL-ROM} \cite{noauthor_c-oral-rom_nodate} 
 & \checkmark &        & \checkmark & \checkmark & \checkmark
 &        &        & \checkmark &        &       
 &        &        &        &        & 
 \\
 \hline
 
\textbf{Common Voice} \cite{noauthor_common_nodate} 
 & \checkmark &        & \checkmark &        &      
 & \checkmark & \checkmark &        &        & \checkmark
 &        & \checkmark &        &        &  
 \\
 \hline
 
\textbf{DEMoS} \cite{parada-cabaleiro_demos_2019,parada-cabaleiro_demos_2020}
 & \checkmark &        & \checkmark &        &      
 &        &        &        & \checkmark &       
 &        &        &        &        &  
 \\
 \hline
 
\textbf{DIA} \cite{vietti_dia-dialogic_2021}
 & \checkmark &        & \checkmark &        & \checkmark
 &        &        &        &        &       
 &        &        &        &        & 
 \\
 \hline
 
\textbf{EMOFILM} \cite{parada-cabaleiro_emofilm_2018}
 & \checkmark &        &        &        & \checkmark
 &        &        &        & \checkmark &       
 &        &        &        & \checkmark & 
 \\
 \hline
 
\textbf{Europarl-ST} \cite{noauthor_europarl_nodate,iranzo-sanchez_europarl-st_2020}
 & \checkmark & \checkmark &        &        & \checkmark
 &        &        & \checkmark &        &       
 &        &        &        &        &  
 \\
 \hline
 
\textbf{IBNC} \cite{noauthor_ibnc_nodate}
 & \checkmark &        &        &        &      
 & \checkmark &        &        &        &       
 &        &        &        &        &  
 \\
 \hline
 
\textbf{Italian Spontaneous Dialogue (Telephony)} \cite{noauthor_italianitaly_nodate}
 &        &        & \checkmark & \checkmark & \checkmark
 & \checkmark &        &        &        &       
 & \checkmark &        &        &        &  
 \\
 \hline
 
\textbf{Italian Kids Speech Recognition Corpus} \cite{noauthor_italian_nodate-12}
 & \checkmark &        &        &        &      
 & \checkmark &        &        &        &       
 &        &        & \checkmark &        &  
 \\
 \hline
 
\textbf{Italian Lifestyle Podcast} \cite{noauthor_italian_nodate-2}
 &        &        &        &        & \checkmark
 & \checkmark &        &        &        &       
 &        &        &        &        &  
 \\
 \hline
 
\textbf{Italian Scripted Monologue} \cite{noauthor_italian_nodate}
 & \checkmark &        &        &        &      
 & \checkmark & \checkmark &        &        &       
 &        &        &        &        &  
 \\
 \hline
 
\textbf{Italian Spontaneous Dialogue Dataset} \cite{noauthor_italian_nodate-1}
 &        &        & \checkmark & \checkmark & \checkmark
 & \checkmark &        &        &        &       
 & \checkmark &        &        &        &  
 \\
 \hline
 
\textbf{KIParla} \cite{noauthor_corpus_nodate,mauri_kiparla_nodate}
 & \checkmark & \checkmark & \checkmark & \checkmark & \checkmark
 &        &        &        &        &       
 &        &        &        &        &  
 \\
 \hline
 
\textbf{LABLITA} \cite{noauthor_lablita_nodate,cresti_lablita_2022}
 & \checkmark &        & \checkmark & \checkmark & \checkmark
 &        &        &        &        &       
 &        &        &        &        &  
 \\
 \hline
 
\textbf{M-AILABS} \cite{celeste_witchzard_m-ailabs-dataset_2024}
 & \checkmark &        &        &        &      
 & \checkmark & \checkmark &        &        &       
 &        &        &        &        &  
 \\
 \hline
 
\textbf{MLS} \cite{pratap_mls_2020,noauthor_mls_nodate}
 & \checkmark &        &        &        &      
 & \checkmark & \checkmark &        &        &       
 &        &        &        &        &  
 \\
 \hline
 
\textbf{MULTEXT Prosodic} \cite{noauthor_multext_nodate}
 & \checkmark &        &        &        &      
 &        &        &        &        &       
 &        &        &        &        &  
 \\
 \hline
 
\textbf{MuST-C} \cite{di_gangi_must-c_2019}
 &        &        &        &        &      
 & \checkmark &        & \checkmark &        &       
 &        &        &        &        &  
 \\
 \hline
 
\textbf{PortMedia} \cite{noauthor_portmedia_nodate}
 &        &        &        &        & \checkmark
 & \checkmark &        & \checkmark &        &       
 &        &        &        &        &  
 \\
 \hline
 
\textbf{VIVALDI} \cite{noauthor_vivaldi_nodate}
 & \checkmark &        & \checkmark & \checkmark & \checkmark
 &        &        &        &        &       
 &        &        &        &        & \checkmark
 \\
 \hline
 
\textbf{VoxPopuli} \cite{noauthor_voxpopuli_2024,wang_voxpopuli_2021}
 & \checkmark &        &        &        & \checkmark
 & \checkmark &        & \checkmark &        &       
 &        &        &        &        &  
 \\
\hline
\end{tabular}
} 
\end{table}
    
\end{landscape}


\section{Challenges and Limitations}
In addition to the challenges highlighted in the introduction regarding datasets created for specific purposes, which, based on the authors' experience, are often difficult to locate or access, this section addresses the challenges and limitations of the datasets to which we have had access.
The aim is to critically evaluate their usability, completeness, and representation in the context of their intended applications, as well as to highlight practical and structural barriers limiting their adoption in various domains.
Addressing these limitations is indeed critical to enhancing their effectiveness and ensuring they meet the needs of researchers and developers.Key challenges include issues related to data scarcity, accessibility, technical constraints, and ethical considerations.

\subsection{Data Scarcity and Representativeness}
One of the most pressing challenges in the creation of spoken Italian datasets is the scarcity of resources that comprehensively represent the linguistic and demographic diversity of Italy. This issue manifests in several ways:
\begin{itemize}[topsep=0pt, partopsep=0pt, itemsep=0pt, parsep=0pt]
\item \textbf{Underrepresentation of Dialects and Minority Languages}.
While datasets like VIVALDI \cite{noauthor_vivaldi_nodate} focus on regional varieties and minority languages, many other datasets prioritize standard Italian, leaving dialects and lesser-spoken languages underrepresented. This creates gaps in linguistic research and limits the development of tools tailored to specific regions or linguistic communities.

\item \textbf{Demographic Imbalance}.
Many datasets disproportionately feature younger, urban speakers, which does not accurately reflect the broader population. For example, older speakers, who often preserve traditional dialects, are frequently underrepresented in large-scale collections, leading to datasets that lack sufficient generational variation.

\item \textbf{Limited Coverage of Specialized Speech Contexts}.
Domains such as pathological speech, learner language, and professional jargon are often overlooked. For instance, while datasets like DEMoS \cite{parada-cabaleiro_demos_2020, parada-cabaleiro_demos_2019} provide emotional speech and the set of datasets offered by FutureBeeAI \cite{noauthor_wake_nodate,noauthor_travel_nodate,noauthor_telecom_nodate,noauthor_retail_nodate,noauthor_healthcare_nodate,noauthor_delivery_nodate,noauthor_bfsi_nodate,noauthor_travel_nodate-1,noauthor_telecom_nodate-1,noauthor_real_nodate-1,noauthor_healthcare_nodate-1,noauthor_bfsi_nodate-1,noauthor_retail_nodate-1,noauthor_delivery_nodate-1,noauthor_italian_nodate-13} are focused on different speaking contexts, there is a relative scarcity of resources addressing medical speech or non-native Italian speakers’ proficiency levels.

\item \textbf{Restricted Access}.
A significant number of existing datasets are only available for a fee or under restrictive licensing agreements, limiting their accessibility to researchers and developers with substantial funding. For example, commercial datasets such as Aurora Project \cite{noauthor_aurora_nodate} Reference] and Defined.ai \cite{noauthor_italian_nodate, noauthor_italian_nodate-1, noauthor_italian_nodate-2} Italian Datasets require significant financial investment, making them inaccessible for smaller research teams or open-source projects. This restricts innovation and reinforces inequalities in data access. In Table \ref{tab:open_d}, we present a list of datasets that are available under different forms of “open access” together with some relevant features.

\item \textbf{Resource Constraints in Data Collection}.
Gathering high-quality, representative data requires significant time and financial investment. For example, dialectal data often necessitate field recordings in remote regions, involving logistical challenges that can limit the scope and size of the datasets.
\end{itemize}

To address these issues, collaborative efforts are needed between academic institutions, government agencies, and private organizations. Initiatives like VoxForge \cite{noauthor_voxforgeorg_nodate} and \textbf{Common Voice} \cite{noauthor_common_nodate}, which leverage crowdsourcing, demonstrate the potential for scaling data collection while enhancing demographic diversity. However, more targeted efforts are required to ensure the inclusion of underrepresented dialects, speech contexts, and demographics, as well as to advocate for open access to key datasets.

\begin{table*}[t]

\centering
\caption{Summary of Publicly Available Italian Speech Datasets}
\small{
\begin{tabular}{|m{2cm}|m{2.5cm}|m{2.5cm}|m{3cm}|m{3.5cm}|}
\hline
\textbf{Dataset} & \textbf{Availability} & \textbf{Size} & \textbf{Type of Speech} & \textbf{Quality Control and Validation} \\ \hline
\textbf{KIParla} \cite{noauthor_corpus_nodate, mauri_kiparla_nodate} & Creative Commons BY-NC-SA 4.0 & 150+ hours & Conversational, spontaneous & Reviewed for accuracy; anonymized \\ \hline
\textbf{MLS (Italian subset)} \cite{noauthor_mls_nodate, pratap_mls_2020} & Open access via OpenSLR & 279.43 hours (Italian part) & Read speech & Transcript-audio alignment reviewed \\ \hline
\textbf{Europarl-ST} \cite{noauthor_europarl_nodate, iranzo-sanchez_europarl-st_2020}& Creative Commons license & $\sim$64.18 hours (Italian part) & Formal parliamentary & Not specified \\ \hline
\textbf{M-AILABS} \cite{celeste_witchzard_m-ailabs-dataset_2024} & Public Domain & $\sim$108 hours (Italian part) & Read speech & Basic alignment validation; variable quality \\ \hline
\textbf{ASR-ItaCSC} \cite{noauthor_asr-itacsc_nodate}& Creative Commons Attribution-NonCommercial-NoDerivatives 4.0 & 10.43 hours & Spontaneous conversation & Internal checks for transcription accuracy \\ \hline
\textbf{Common Voice} \cite{noauthor_common_nodate} & CC0 Public Domain Dedication & $\sim$1,000 hours (Italian) & Read speech & Community-driven validation \\ \hline
\textbf{VIVALDI} \cite{noauthor_vivaldi_nodate}& Open for research and education & 14+ regions (size varies) & Spontaneous, elicited & Validated by linguistic experts \\ \hline
\textbf{MuST-C} \cite{di_gangi_must-c_2019} & Creative Commons (temporarily suspended) & $\sim$385 hours (Italian part) & TED talks, translations & Standard transcription and alignment processes \\ \hline
\textbf{GEMMA Project} \cite{benedetto_consonant_2020}& Available upon request. Licensed under CC-BY-NC-ND & n.a. & Isolated syllables focusing on consonant gemination & n.a. \\ \hline
\end{tabular}}
\label{tab:open_d}
\end{table*}

\subsection{Technical Challenges}
Technical challenges in creating and utilizing spoken Italian datasets often stem from the complexities of speech data itself and the diverse contexts in which it is collected. These challenges impact the quality, consistency, and usability of datasets across various applications. Key technical issues include:
\begin{itemize}[topsep=0pt, partopsep=0pt, itemsep=0pt, parsep=0pt]
\item \textbf{Audio Quality and Background Noise}
Many datasets, especially those collected in uncontrolled environments, suffer from low audio quality due to background noise, poor recording equipment, or environmental interference. For example, field recordings in datasets like VIVALDI \cite{noauthor_vivaldi_nodate} often include ambient noise, which, while sometimes valuable for realism, can hinder the development of automatic speech recognition (ASR) systems that require clean audio for training.

\item \textbf{Inconsistent Recording Conditions}
Variability in recording setups—such as microphone quality, distance from speakers, and recording environments—leads to inconsistent audio quality. Datasets like Aurora Project \cite{noauthor_aurora_nodate} address this challenge by using standardized protocols, but achieving such consistency across large-scale datasets remains difficult.

\item \textbf{Large File Sizes and Processing Overheads}
Speech data, especially high-quality recordings, results in large file sizes, creating challenges in storage, processing, and distribution. Researchers working with extensive datasets, such as MLS (Multilingual Large Scale Dataset) \cite{pratap_mls_2020, noauthor_mls_nodate}, often need substantial computational resources for annotation, segmentation, and modeling.

\item \textbf{Alignment and Segmentation Errors}
The accuracy of time-aligned transcriptions and segmentation is critical for many speech applications. Automated tools such as WebMAUS \cite{noauthor_webmaus_nodate} or Praat \cite{noauthor_praat_nodate} reduce manual labor but are not immune to alignment errors, particularly in datasets with spontaneous or noisy speech, such as KIParla \cite{mauri_kiparla_nodate, noauthor_corpus_nodate}.

\item \textbf{Annotation Standardization Across Datasets}
Inconsistent annotation schemas between datasets complicate cross-dataset comparisons and interoperability. For instance, while one dataset might use orthographic transcription, another may employ phonetic or prosodic annotations, making integration for large-scale studies challenging without extensive preprocessing.

\item \textbf{Multimodal Integration Challenges}
Increasingly, datasets combine audio with additional modalities, such as text or visual cues. While multimodal corpora like \textbf{VoxPopuli} \cite{noauthor_voxpopuli_2024, wang_voxpopuli_2021} offer immense potential for research, they also require sophisticated tools for synchronizing and analyzing diverse data types.

\item \textbf{Dialectal and Linguistic Variability}
Capturing the nuances of Italian dialects and linguistic diversity adds complexity to dataset creation. Phonological differences, code-switching, and regional variations can introduce inconsistencies that are difficult to standardize without specialized linguistic expertise, as seen in datasets like VIVALDI \cite{noauthor_vivaldi_nodate}.

\item \textbf{Model Robustness to Real-World Variability}
Speech datasets collected under ideal conditions often fail to prepare models for real-world deployment, where variability in accents, speaking styles, and environments is the norm. Datasets like \textbf{Common Voice} \cite{noauthor_common_nodate}, which include speaker diversity, are steps toward addressing this challenge, but gaps remain in simulating real-world scenarios effectively.
\end{itemize}

\subsection{Ethical Considerations}
The creation and use of spoken Italian datasets raise several ethical concerns that must be addressed to ensure responsible data collection, annotation, and application. These concerns primarily revolve around issues of privacy, consent, and fairness.
\begin{itemize}[topsep=0pt, partopsep=0pt, itemsep=0pt, parsep=0pt]
\item \textbf{Privacy and Data Protection}
Recording speech inherently involves capturing personal and identifiable information. Without proper safeguards, datasets risk exposing sensitive data such as speaker identities or private conversations. For instance, datasets collected via telephony, like the Italian Spontaneous Dialogue Telephony Speech Dataset \cite{noauthor_italianitaly_nodate}, must anonymize recordings to protect speaker privacy. Compliance with regulations such as the EU’s General Data Protection Regulation (GDPR) \cite{noauthor_gdpr_nodate} is essential to ensure lawful data collection and use.

\item \textbf{Informed Consent}
Ethical data collection requires obtaining explicit consent from participants, ensuring they understand how their data will be used, stored, and shared. Field recordings for datasets such as VIVALDI \cite{noauthor_vivaldi_nodate}, which involve speakers of minority languages, highlight the importance of clear communication with participants, especially in remote or less formally regulated contexts.

\item \textbf{Fair Representation}
Ensuring demographic diversity in datasets is critical for avoiding biases in speech technologies. Datasets skewed toward certain age groups, genders, or accents can perpetuate inequality in applications like ASR or voice assistants. For instance, datasets such as \textbf{Common Voice} \cite{noauthor_common_nodate}, which emphasize inclusivity, set a benchmark for equitable data representation. However, many datasets still fall short, disproportionately representing urban or standard Italian speakers while neglecting rural or dialectal varieties.

\item \textbf{Ethical Use of Data}
Datasets designed for one purpose may be repurposed in ways that conflict with the original ethical agreements. For example, speech datasets could be misused to develop surveillance systems or deepfake technologies without the consent of the original contributors. Ensuring proper licensing and monitoring of dataset use is crucial to mitigate these risks.

\item \textbf{Cultural Sensitivity}
In projects documenting dialects and minority languages, such as VIVALDI \cite{noauthor_vivaldi_nodate}, ethical considerations extend to the cultural significance of the data. Researchers must engage with local communities to ensure that their linguistic heritage is documented respectfully and that communities retain agency over how the data is used and shared.
\end{itemize}
Addressing these ethical challenges requires a multi-faceted approach, including robust anonymization protocols, transparent consent procedures, inclusive dataset design, and culturally sensitive practices. By adhering to ethical standards, researchers and developers can ensure that spoken Italian datasets contribute positively to both technological innovation and societal well-being.

\section{Future Directions and Recommendations}
To address the current limitations of spoken Italian datasets, future efforts should focus on expanding demographic and linguistic diversity, including underrepresented dialects, minority languages, and diverse speaker groups. Leveraging advancements in AI-driven annotation and multimodal data integration can reduce costs and improve dataset quality.
Promoting open access and fostering collaborations between academia, industry, and communities will ensure broader availability and encourage innovation. Standardizing annotation and metadata practices across datasets is also essential for enabling large-scale comparisons and improving interoperability.
Ethical considerations, such as ensuring privacy and cultural sensitivity, must remain a priority to build trust and prevent misuse. Interdisciplinary collaboration between linguists, technologists, and cultural experts can further maximize the impact of these datasets in both research and real-world applications.
By addressing these challenges and opportunities, spoken Italian datasets can continue driving advancements in linguistics and technology while preserving Italy’s rich linguistic heritage.

\section{Conclusion}

This work has presented a comprehensive survey of 66 spoken Italian datasets, examining their types, collection and annotation methodologies, as well as current limitations. In the introductory section (Section 1), we highlighted the gap between the resources available for Italian and those for major world languages such as English or Mandarin. Subsequently, we have proposed a classification of the datasets according to speech type (spontaneous, conversational, read, monologic), source and context (broadcast media, telephone, etc.), and demographic or linguistic features (dialects, sociolinguistic variation, multilingual settings).

In the successive Section, we focused on data collection and annotation strategies, underscoring the importance of standardized protocols, automated tools, and validation procedures (e.g., inter-annotator agreement) to ensure high-quality data. Section 4 has discussed key applications, ranging from automatic speech recognition (ASR) and text-to-speech (TTS) to emotion analysis and language learning, as well as cultural and heritage preservation.

We then addressed the main challenges and limitations: limited coverage of dialects and minority languages, restricted access and licensing issues, demographic imbalances, and ethical considerations related to privacy. Finally, Section 6 outlined several future recommendations, including a call for standardized annotation formats, closer collaboration between academic and industrial stakeholders, and the need for crowdsourced initiatives to broaden dialectal and linguistic coverage.

Based on these findings, the important actions to be carried on have been presented, such as
expanding datasets to include underrepresented dialects and regional varieties, potentially by partnering with local communities for field recordings;
promoting open access and revise data-sharing policies to enhance research and innovation, particularly for educational contexts and smaller enterprises; developing and adopting consistent annotation and validation standards (covering demographic, sociolinguistic, and prosodic metadata) to improve dataset interoperability; fostering interdisciplinary research and partnerships with industrial, cultural, and international academic institutions, aiming to advance Italian speech technologies on a global scale.

To support the exploration and development of spoken Italian datasets, this survey highlights a curated set of 66 resources, encompassing diverse speech types, sources, and linguistic features. While this paper discusses selected examples to illustrate the challenges, methodologies, and applications in the field, the full collection is made publicly available for further reference and research. The complete dataset repository, hosted on \href{https://github.com/marco-giordano/spoken-italian-datasets}{GitHub}  and archived on Zenodo (DOI: 10.5281/zenodo.14246196), aims to facilitate collaborative efforts and inspire future advancements in Italian linguistic research and speech technology.

\bibliographystyle{IEEEtran}
\bibliography{References.bib}

\begin{IEEEbiography}
[{\includegraphics[width=1in,height=1.35in,clip,keepaspectratio]{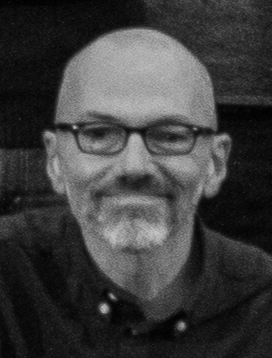}}]
{Marco Giordano} is a Full Professor in Musical Acoustics and Perception at the Conservatory of Music of L'Aquila, Italy and currently PhD student in ICT and Telecommunications Engineering at the DISIM dpt. of the University of L'Aquila. He has a Laurea degree in Electronic Engineering from the University of Roma. His main research activities are focused on digital signal processing, musical acoustics and perception, digital education and deep neural models applied to computational paralinguistics and sound recognition.
\end{IEEEbiography}

\begin{IEEEbiography}
[{\includegraphics[width=1in,height=1.35in,clip,keepaspectratio]{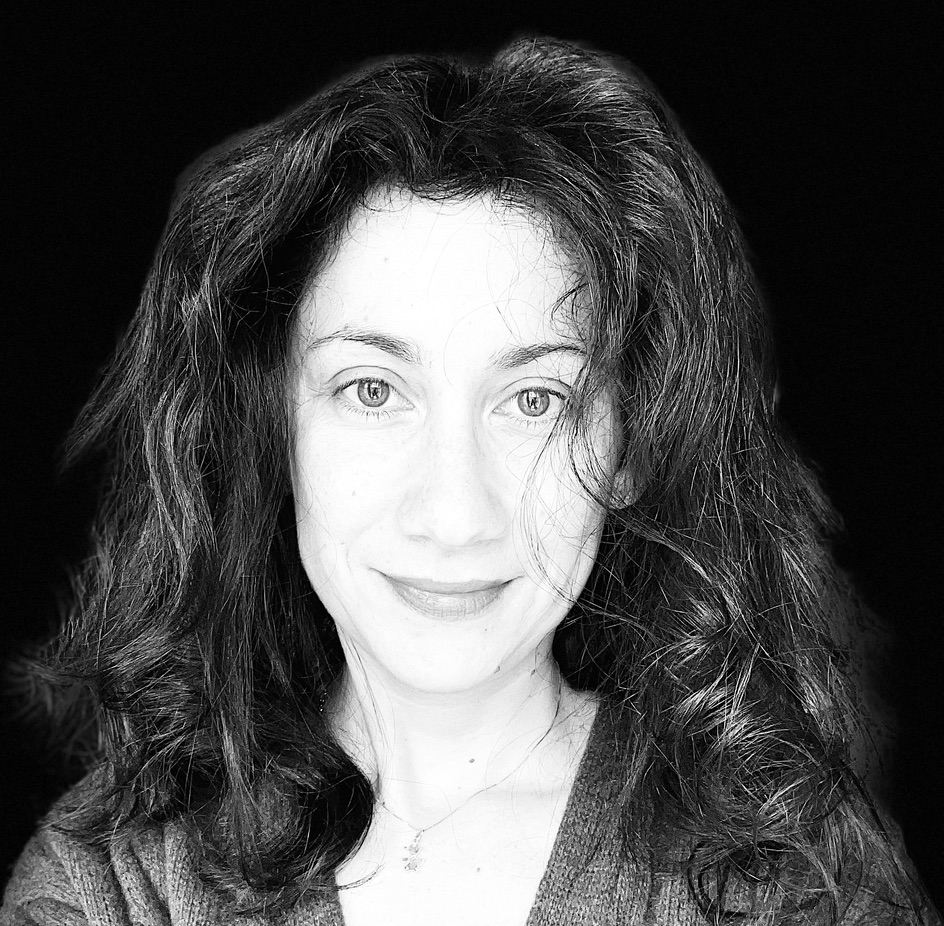}}]
{Claudia Rinaldi} is a Researcher in Signal Processing at CNIT (National Inter-University Consortium for Telecommunications) and Adjunct Professor at the Department of Information Engineering, Computer  Science  and  Mathematics  of  the  University of L’Aquila, Italy. She received the Laurea degree (cum laude) and Ph.D. degree in Electronic Engineering from the University of L'Aquila, Italy, in 2005 and 2009, respectively. She also got a master degree in Trumpet and bachelor degree in Electronic Music at the Conservatory of Music of L'Aquila in 2006 and 2013 respectively. 

Her main research activities are focused on digital signal processing algorithms and more in general on the use of technology in artistic fields. Moreover, she is concerned with design, modeling and optimization of communication algorithms with particular emphasis on the physical layer and software defined radio for the development of transmission systems responding to cognitive radios paradigms.
\end{IEEEbiography}

\vfill
\EOD
\end{document}